\documentclass[letterpaper, 10 pt, journal, twoside]{ieeetran}

\IEEEoverridecommandlockouts                              

\usepackage{graphics} 
\usepackage{epsfig} 
\usepackage{times} 
\usepackage{amssymb}  
\usepackage{bm}
\usepackage{xcolor}
\usepackage{cite}
\usepackage{array}
\usepackage{booktabs}
\usepackage{multirow}

\usepackage{gensymb,soul}
\usepackage{ulem} 
\usepackage{hyperref}
\hypersetup{pdfstartview=FitH,
            colorlinks=true,
            linkcolor=red,
            anchorcolor=blue,
            citecolor=black
            }
\usepackage{amsthm}
\usepackage{float}
\usepackage{booktabs}
\usepackage{multirow}
\usepackage{makecell}
\usepackage{enumerate}
\usepackage{leftindex}
\usepackage{mathtools}
\usepackage{mathrsfs}
\usepackage[utf8]{inputenc}
\usepackage[caption=false,font=footnotesize]{subfig}
\usepackage{graphicx}
\usepackage{pifont}
\usepackage{soul}
\usepackage[linesnumbered,ruled]{algorithm2e}
\usepackage{bbding}

\newcounter{RNum}
\renewcommand{\theRNum}{\arabic{RNum}}
\theoremstyle{plain}

\newtheorem{theorem}{Theorem}

\newcommand{\Remark}{\noindent\textbf{Remark}~\refstepcounter{RNum}\textbf{\theRNum}: }
\newcommand{\NoOne}[1]{\textcolor{red}{#1}}
\newcommand{\NoTwo}[1]{\textcolor{green}{#1}}
\newcommand{\NoThree}[1]{\textcolor{blue}{#1}}
\makeatletter 
\pretocmd\@bibitem{\color{black}\csname keycolor#1\endcsname}{}{\fail}
\newcommand\citecolor[1]{\@namedef{keycolor#1}{\color{blue}}}
\makeatother
\newcommand{\re}{\mathbb{R}}
\newcommand{\qm}{\mbox{Qmod}}

\soulregister\NoOne7
\soulregister\NoTwo7
\soulregister\NoThree7
\soulregister\Remark7

\begin{document}

\title{
Collision-Free Trajectory Optimization in Cluttered Environments Using Sums-of-Squares Programming
}

\author{Yulin Li$^{1}$, Chunxin Zheng$^{2}$, Kai Chen$^{2}$, Yusen Xie$^{2}$, Xindong Tang$^{3}$,\\ Michael Yu Wang$^{4}$, \textit{Fellow, IEEE,} and Jun Ma$^{1}$ 

\thanks{$^{1}$Yulin Li and Jun Ma are with the Division of Emerging Interdisciplinary Areas, The Hong Kong University of Science and Technology, Hong Kong SAR, China {\tt\footnotesize yline@connect.ust.hk; jun.ma@ust.hk}}
\thanks{$^{2}$Chunxin Zheng, Kai Chen, and Yusen Xie are with the Robotics and Autonomous Systems Thrust, The Hong Kong University of Science and Technology (Guangzhou), Guangzhou, China {\tt\footnotesize czheng739@connect.hkust-gz.edu.cn}}
\thanks{$^{3}$Xindong Tang is with the Department of Mathematics, Hong Kong Baptist University, Hong Kong SAR, China {\tt\footnotesize xdtang@hkbu.edu.hk}}%
\thanks{$^{4}$Michael Yu Wang is with the School of Engineering, Great Bay University, China {\tt\footnotesize mywang@gbu.edu.cn}}%
\thanks{\,\,\,Yulin Li and Chunxin Zheng contributed equally to this work.}
}
 


\maketitle


\begin{abstract}

In this work, we propose a trajectory optimization approach for robot navigation in cluttered 3D environments. 
We represent the robot's geometry as a semialgebraic set defined by polynomial inequalities such that robots with general shapes can be suitably characterized. To address the robot navigation task in obstacle-dense environments, we exploit the free space directly to construct a sequence of free regions, and allocate each waypoint on the trajectory to a specific region. Then, we incorporate a uniform scaling factor for each free region, and formulate a Sums-of-Squares (SOS) optimization problem that renders the containment relationship between the robot and the free space computationally tractable.
The SOS optimization problem is further reformulated to a semidefinite program (SDP), and the collision-free constraints are shown to be equivalent to limiting the scaling factor along the entire trajectory. 
In this context, the robot at a specific configuration is tailored to stay within the free region.
Next, to solve the trajectory optimization problem with the proposed safety constraints, we derive a guiding direction for updating the robot configuration to decrease the minimum scaling factor by calculating the gradient of the Lagrangian at the primal-dual optimum of the SDP. As a result, this seamlessly facilitates the use of gradient-based methods in efficient solving of the trajectory optimization problem. 
Through a series of simulations and real-world experiments, the proposed trajectory optimization approach is validated in various challenging scenarios, and the results demonstrate its effectiveness in generating collision-free trajectories in dense and intricate environments populated with obstacles. Our code is available at: \url{https://github.com/lyl00/minimum_scaling_free_region}.

\end{abstract}

\section{Introduction} \label{sec:intro}

 Over the past decade, substantial progress has been made in trajectory optimization for a diverse range of applications in robotics\cite{9905530,6631299,9196996}. 
 At its core, it aims to optimize for a collision-free trajectory that adheres to motion constraints while minimizing a task-specific cost function. 
 Recent accomplishments in numerical optimization have facilitated effective solutions for nonlinear programs with various general constraints and objectives. However, 
the inherent complexity and nonconvexity of the collision-free configuration space still 
 pose significant challenges in effectively modeling safety constraints, particularly when confronted with intricate and densely populated obstacle layouts. Moreover, accurately formulating collision avoidance constraints often necessitates considering the geometry of the robot and its surrounding environment, which is difficult to capture explicitly amidst complex robot and obstacle configurations \cite{10417140}. 
 
 
 In this paper, we present a trajectory optimization approach that generates collision-free motions for robots with general geometries in intricate and cluttered environments. Specifically, it is achieved by enforcing the robots to stay within the collision-free workspace over the optimization horizon.
 We begin by approximating the free space by decomposing it into a series of polytopic regions and constructing a graph to capture how these regions interconnect. Subsequently, we search for a free region sequence and generate a path from the starting point to the destination that traverses through these free regions with the least cumulative distance. We formulate a Sums-of-Squares (SOS) optimization problem to solve for the minimum scaling factor that certifies the containment relationship between the robot and the scaled free region, considering its current configuration. The collision-free constraints are then enforced by limiting the scaling factor along the trajectory and the geometries of the robots are represented by sets of polynomial inequalities, thereby facilitating the application of this methodology to robots with general shapes in a unified framework. Moreover, the gradient of the scaling factor with respect to the robot configuration is subsequently approximated, which is effectively integrated with the augmented Lagrangian 
 iterative linear quadratic regulator (AL-iLQR)\cite{8967788}, yielding a robust and efficient pipeline that steers the trajectory towards safety and optimality with rapid convergence. Lastly, through extensive evaluations in a variety of challenging scenarios, our proposed framework demonstrates its feasibility and efficacy in generating safe and efficient trajectories, and the capability for flexible self-pose adjustment adapting to complex and cluttered environments is attained.

\section{Related Works}
As an indispensable ingredient in trajectory optimization, various approaches have been explored in the literature to formulate collision avoidance constraints in cluttered environments. 
One category involves exploiting the distance information between robots and obstacles. In \cite{constraint-DDP}, distance margins between the robot and obstacles are explicitly represented and enforced as state constraints. 
However, the distance between two general shapes is inherently implicit, inevitably leading to oversimplified collision geometries and conservative maneuvers. In \cite{9812334}, collision avoidance between polytopic robots and obstacles is considered. It formulates a quadratic programming (QP) program that solves for the minimum distance between two polytopic sets. By examining the dual perspective of the QP, it derives explicit control barrier function
(CBF) constraints and solves the nonlinear nonconvex optimization problem directly using IPOPT\cite{wachter2006implementation}. This method
introduces numerous additional dual variables based on the optimization step size. Another recent work \cite{10160716} defines cone representations for several common geometry primitives and constructs a minimum scaling problem for every pair of collision primitives for collision checking. However, this approach requires prior knowledge of obstacle geometries and restricts the robot's shape to predefined primitives. Besides, different primitive pairs necessitate distinct optimization problems, which hinders their implementation in environments with complex obstacle configurations. Additionally, these distance-based methods rely on formulating distance constraints for each robot-obstacle pair, which significantly increases the computational burden in cluttered environments.
Alternatively, some works have attempted to approximate obstacle-free workspaces with parametric free regions, such as polytopic free regions in Euclidean space \cite{7839930} and elliptical safe corridors parameterized along a path \cite{arrizabalaga2024differentiablecollisionfreeparametriccorridors}. In \cite{7839930}, the robot is constrained to stay within the polytopic free regions. However, this method assumes a point robot model due to the intractability of the containment relationship between sets with general geometries. In \cite{doi:10.1126/scirobotics.adf7843}, the collision-free trajectory is generated in the high-dimensional configuration free space for robots with specific kinematic chains, whose convex decomposition is obtained using the alternating optimization framework proposed in \cite{doi:10.1177/02783649231201437}. While this approach represents an innovative effort to map high-dimensional, obstacle-free spaces, the process of achieving such decomposition proves to be notably time-intensive.

Numerous nonlinear programming (NLP) solvers have been developed to deal with trajectory optimization problems with nonlinear system dynamics\cite{9905530,wachter2006implementation}. Among these, sequential convex programming (SCP) stands out as a robust strategy, which effectively approximates solutions through a series of convex subproblems. For instance, sequential quadratic programming (SQP) serves as the backbone algorithm in prominent tool such as TrajOpt\cite{doi:10.1177/0278364914528132}.
Despite these advancements, the computational complexity has historically been a barrier to the broader application of SQP methods.
In contrast, indirect methods like the iterative linear quadratic regulator (iLQR), offer advantages over these direct approaches, including reduced memory usage and enhanced computational speed. Extensions have been developed based on the barrier method\cite{9766194},  distributed optimization \cite{9861337}, and augmented Lagrangian techniques \cite{8206457} to overcome the limitations of DDP-based methods in accommodating various constraints. Recently, ALTRO stands out as a significant advancement\cite{8967788}. By incorporating iLQR within an augmented Lagrangian framework, ALTRO addresses both state and input constraints, demonstrating rapid convergence, numerical stability, and outperforming several benchmark solvers across a range of motion-planning challenges.


\section{PROBLEM STATEMENT}
We list relevant notations in Table \ref{tab:some_definition} for convenience. This work aims to generate collision-free trajectories for robots with specific geometries in obstacle-dense 3D environments. Specifically, we approximately decompose the obstacle-free workspace $\mathcal{F}$ into $N$ overlapping polytopic regions $\mathcal{Q}_i$ with $i=1,2,...,N$, i.e., $
\cup_{i=1:N}\mathcal{Q}_i = \mathcal{Q}\approx\mathcal{F}$. Then, $\boldsymbol{q}$ is considered as a safe configuration if $\mathcal{W}\left(\boldsymbol{q}\right)\subseteq \mathcal{Q}$. Given the above definitions, our goal is to solve for the trajectory $\mathcal{X} = \left\{\boldsymbol{q}_0, \boldsymbol{q}_1,...,\boldsymbol{q}_T\right\}$ and $U = \left\{\boldsymbol{u}_0, \boldsymbol{u}_1,...,\boldsymbol{u}_{T-1}\right\}$, which is the solution of the following nonlinear optimization problem over the horizon $T$:
\begin{align}\label{eqn:to}
    \displaystyle  \operatorname*{minimize}_{(\boldsymbol{q}_{\tau},\boldsymbol{u}_{\tau})\in\re^{n}\times\re^{m}}\quad & \phi_T(\boldsymbol{q}_{T})+\sum_{\tau=0}^{T-1}J_\tau\big(\boldsymbol{q}_{\tau},\boldsymbol{u}_{\tau}\big)\notag\\ 
    \operatorname*{subject\ to}\quad \,\, & \boldsymbol{q}_{\tau+1}=f\big(\boldsymbol{q}_{\tau},\boldsymbol{u}_{\tau}\big),\notag\\    & h_{\tau}(\boldsymbol{q}_{\tau},\boldsymbol{u}_{\tau}) = 0,\notag\\ &g_{\tau}(\boldsymbol{q}_{\tau},\boldsymbol{u}_{\tau}) \leq 0,\yesnumber\\
    &\quad\quad \tau = 0,1,\ldots,T-1\notag\\
    & \mathcal{W}_\mathcal{B}\left(\boldsymbol{q}_{\tau}\right)\subseteq \mathcal{Q},\notag\\
    &\quad\quad \tau = 0,1,\ldots,T\notag\\
    &\boldsymbol{q}_0 = \boldsymbol{q}_{s}.\notag
\end{align}
In the above, the subscript $\tau$ denotes the corresponding values at that time stamp;   
$J_\tau\big(\boldsymbol{q}_{\tau},\boldsymbol{u}_{\tau}\big)$ and $\phi_T$ represent the intermediate and terminal costs, respectively; $f\big(\boldsymbol{q}_{\tau},\boldsymbol{u}_{\tau}\big)$ is the discrete dynamic function; $h_{\tau}(\boldsymbol{q}_{\tau},\boldsymbol{u}_{\tau})$ and $g_{\tau}(\boldsymbol{q}_{\tau},\boldsymbol{u}_{\tau})$ denote general equality and inequality state-input constraints. Importantly, the constraint $\mathcal{W}_\mathcal{B}\left(\boldsymbol{q}_{\tau}\right)\subseteq \mathcal{Q}$ states that the space occupied by the robot $\mathcal{B}$ at each time stamp is contained in the sets of free regions, thereby ensuring geometry-aware collision-free maneuvers along the whole trajectory.  
Details of the problem formulation as well as the developed 
methodology to solve this problem are elaborated in the following section.

\begin{table}[t]
    \centering
    \caption{NOMENCLATURE}
    \begin{tabular}{ cc }
    \toprule[1pt]
    \textbf{Symbols} & \textbf{Descriptions}\\
    \midrule
    $\boldsymbol{q}\in\re^n$ & $n$-dimensional robot configuration\\
    $\boldsymbol{u}\in\re^m$ & $m$-dimensional control input\\
    $\mathcal{S}\subset \re^3$ & 3D Euclidean workspace\\ 
    $\mathcal{W}\left(\boldsymbol{o_i}\right)\subset \re^3$ & Space occupied by the $i$th obstacle \\
    $\mathcal{F}$ & Obstacle-free workspace, i.e., $\mathcal{S}\setminus \cup_{o_i\in o}\mathcal{W}\left(\boldsymbol{o_i}\right)$\\
    $\mathcal{W}_\mathcal{B}\left(\boldsymbol{q}\right)\subset \re^3$ & Space occupied by robot $\mathcal{B}$ at configuration $\boldsymbol{q}$\\
    $\boldsymbol{q}_s$ & Start configuration\\
    $\boldsymbol{q}_g$ & Goal configuration\\
    $C(\mathcal{Q})$ & Geometrical center of a convex shape $\mathcal{Q}$\\
    \bottomrule[1pt]
    \end{tabular}
    \label{tab:some_definition}
\end{table}

\section{Methodology}

\subsection{Waypoints Allocation on Graph of Free Regions}\label{sec:shortest_path}
\begin{figure}[t]	
	\centering
        \vspace{3pt}
	\includegraphics[trim=0cm 0cm 0cm 0.0cm, clip,width=0.90\linewidth]{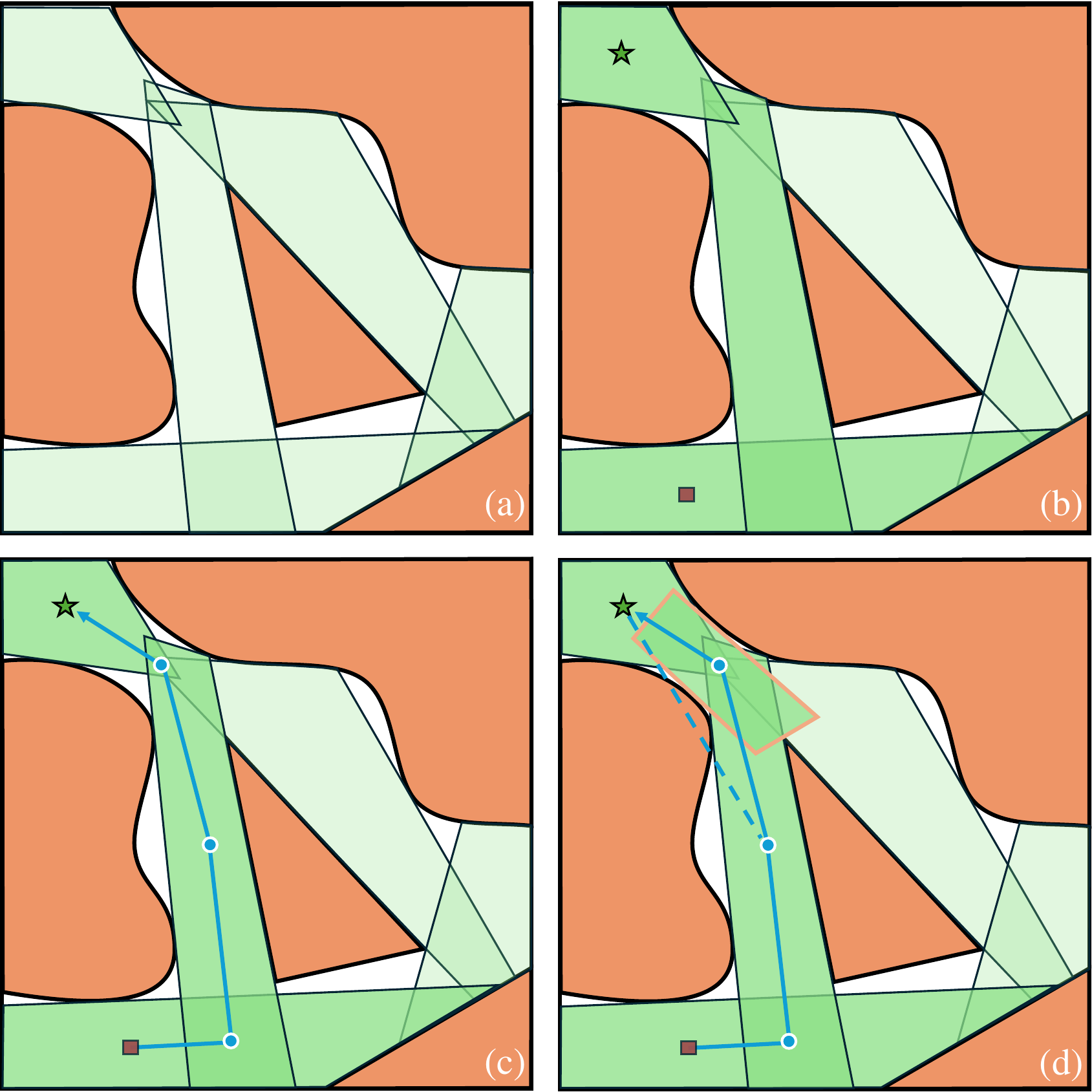}
	\setlength{\abovecaptionskip}{-0pt} 
	\caption
	{Illustration of the waypoints allocation process on the constructed graph of free regions $\mathcal{G}$. (a) The free space $\mathcal{F}$ is decomposed into series of overlapping polytopic regions $\mathcal{Q}$ (the light green polytopic regions). (b) A sequence of the free regions connecting $\boldsymbol{q}_s$ (the red square) and $\boldsymbol{q}_g$ (the green star) with minimal path cost is then searched and stored in $\mathbb{P}^{*}_{v}$ (emphasized in dark green). (c) Waypoints are allocated to specific free regions along the path connecting the edge between each pair of regions in $\mathbb{P}^{*}_{v}$. (d) Extra regions are generated at the geometrical center of particular overlapped area and the waypoints in each appended region is reallocated for smooth transition.}
    \label{fig:shortest_path}
\end{figure}

As shown in Fig. \ref{fig:shortest_path}(a), in an environment with random obstacles, we first generate a family of polytopic free regions by sampling points in $\mathcal{F}$ and using the decomposition algorithm in \cite{7839930} till the portion of remaining free grids is below a threshold (1\%-5\%). Then, we construct an undirected graph $\mathcal{G}:=(\mathcal{V},\mathcal{E})$ from the generated free regions $\mathcal{Q}$. Specifically, each region $\mathcal{Q}_i$ represents a vertex (node) $v_i \in \mathcal{V}$. Two nodes are connected with an edge $e_{i,j}$ if regions $i$ and $j$
intersect sufficiently. Denote the overlapping area as $\mathcal{Q}_{i,j}$, and 
we represent each edge $e_{i,j}$ as the line segments that connect $\mathcal{Q}_i$, $\mathcal{Q}_{i,j}$, and $\mathcal{Q}_j$ sequentially through their geometrical centers, and edge cost $\ell_{e_{i,j}}$ is the length of the straight lines.

Subsequently, we search for a sequence of free regions $\mathbb{P}^{*}_{v}$ connecting $\boldsymbol{q}_s$ and $\boldsymbol{q}_g$ with the minimal cumulative edge length shown as the dark green regions in Fig. \ref{fig:shortest_path}(b)) by applying Dijkstra algorithm on $\mathcal{G}$:
\begin{equation*}
    \mathbb{P}^{*}_{v} = \left\{\{\boldsymbol{q}_s, \mathcal{Q}^{*}_0,\mathcal{Q}^{*}_1,\ldots,\mathcal{Q}^{*}_p,\boldsymbol{q}_g\} \left|\begin{array}{c}
    \mathcal{W}\left(\boldsymbol{q}_s\right)\subseteq \mathcal{Q}^{*}_0,\,\\ \mathcal{W}\left(\boldsymbol{q}_g\right)\subseteq \mathcal{Q}^{*}_p
    \end{array}
    \right.\right\}.
\end{equation*}
Two additional edges $e^*_{s,0}=(\boldsymbol{q}_s, v^*_0)$ and $e^*_{p,g}=(v^*_p,\boldsymbol{q}_g)$ as the line segments connecting the start position to $C(\mathcal{Q}^{*}_0)$ and $C(\mathcal{Q}^{*}_p)$ to the goal position, respectively. Then, the edge sequence concatenating each vertex in $\mathbb{P}^{*}_{v}$ can be extracted accordingly:
\begin{equation*}
    \mathbb{P}^{*}_{e} = \{e^*_{s,0}, e^{*}_{0,1},e^{*}_{1,2},\ldots,e^{*}_{p-1,p},e^*_{p,g}\}.
\end{equation*}

For a given optimization horizon $T$, we discretize the path along the edge sequence $\mathbb{P}^{*}_{e}$ by assuming a constant linear velocity to obtain $T+1$ discrete waypoints. Then, we allocate each waypoint to the free region containing that point and the points inside an overlapping area will always be assigned to the latter one:
\begin{equation}\label{eq:safety_constraints}
\mathcal{W}_\mathcal{B}\left(\boldsymbol{q}_{\tau}\right)\subseteq \mathcal{Q}_{\tau}\quad \forall\, \tau = 0,1,\ldots,T.
\end{equation}

Additionally, as indicated in Fig. \ref{fig:shortest_path}(d), we generate an extra polytopic region at the geometrical center of an overlapped area $\mathcal{Q}_{i,j}$. This mechanism will be triggered if there exist points on the line segment connecting $C(\mathcal{Q}_i)$ and $C(\mathcal{Q}_j)$ that do not belong to either free region, and the points within the appended region will be reassigned. This intuitive strategy provides the trajectory optimization module with improved quality of free regions, thus facilitating smooth transitions in areas with sharp turns.

\Remark{We can determine the intersection by stacking the linear
constraints defining the two polytopic sets. Consequently, the existence of overlap can be straightforwardly identified during the process of constructing the graph.} 

\subsection{Safety Constraints Formulation with SOS Programming}\label{sec:safety_constraints_formulation}
\begin{figure}[t]	
	\centering
        \vspace{3pt}
	\includegraphics[trim=0cm 0cm 0cm 0cm,clip,width=0.82\linewidth]{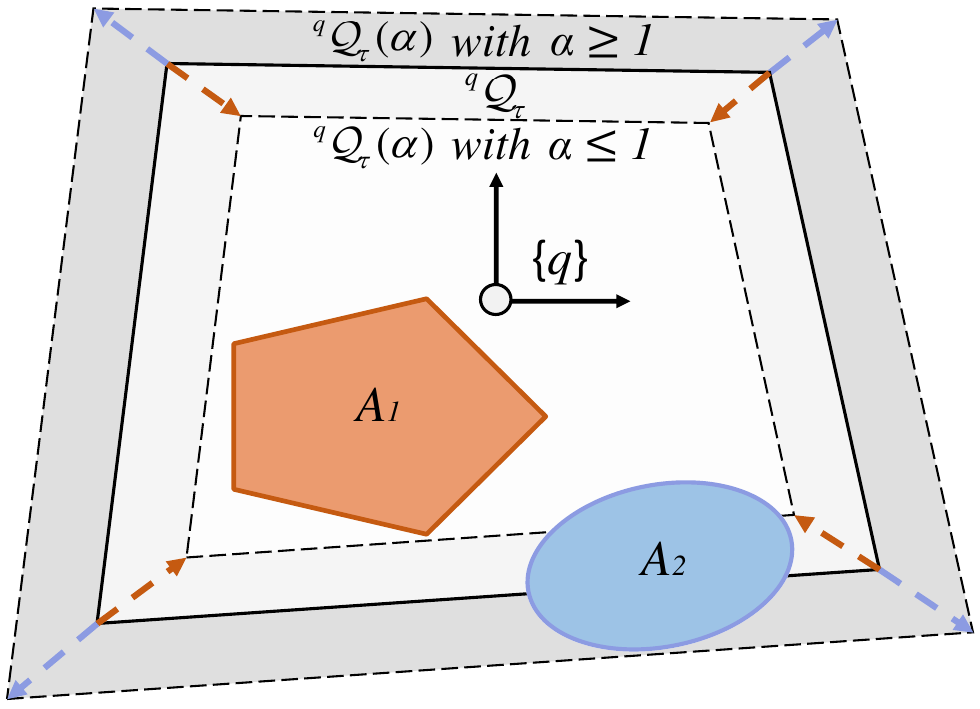}
	\setlength{\abovecaptionskip}{-0pt} 
	\caption
	{Geometrical interpretation of the proposed minimum scaling problem. We aim to find the minimum scaling factor $\alpha$ for the polytopic free region ${}^q\mathcal{Q}_{\tau}$ such that the robots $A_\textit{1}$  ($
 \alpha\leq1$) and $A_\textit{2}$ ($\alpha \geq 1$) are contained in the scaled region.}
	\label{fig:ill_scaling}
\end{figure}

In this section, we introduce the transformation of the safety condition (\ref{eq:safety_constraints}) into an SOS programming problem. Specifically, as illustrated in Fig. \ref{fig:ill_scaling}, we define $\mathcal{Q}_{\tau}(\alpha)$ as the uniformly scaled polytopic free region with a nonnegative scaling factor $\alpha$. Then, an optimization problem based on nonnegative polynomials cone is derived (Equation (\ref{eq:safety_constraint_cone_programming})) and solved to optimal via its SOS relaxation (Equations (\ref{eq:minimal_scaling}-\ref{eq:minimal_scaling_sdp})). It solves for the minimal scaling factor $\alpha$ such that the containment relationship between the robot and the scaled free region can be certified. In this sense, safety can be guaranteed by enforcing the scaling factor $\alpha\leq1$ along the trajectory without loss of generation. It is worthwhile to note that the minimum scaling factor $\alpha$ is unique and bounded since the scaled free region can cover the whole Euclidean space $\re^3$ as $\alpha$ goes to infinity. 
\subsubsection{Set Representation with Polynomial Functions}
We first introduce the necessary basis for establishing the containment relationship between two semialgebraic sets defined with polynomial inequalities.

Given a vector $\boldsymbol{x} = (x_1,x_2\ldots,x_z)$ in $\re^z$ concatenated column-wisely, we denote $\re[\boldsymbol{x}]$ as the space of all real-coefficient multivariate polynomial functions in $\boldsymbol{x}$, and $\re[\boldsymbol{x}]_d$ is the subset of $\re[\boldsymbol{x}]$ containing polynomials with degrees no greater than $d$. 
We define $\boldsymbol{x}^{\beta}:=x_1^{\beta_1}x_2^{\beta_2}\dots x_z^{\beta_z
}$ a \textbf{\textit{monomial}} in $\boldsymbol{x}$ for the tuple of nonnegative integers $\beta:=(\beta_1,\beta_2\dots,\beta_z)$.
Then, a multivariate polynomial function $f(\boldsymbol{x})\in \re[\boldsymbol{x}]$ can be uniquely expressed with a linear combination of all its monomials:
$$f(\boldsymbol{x}) = {\rm coef}(f)^{\top} [\boldsymbol{x}]_{\deg(f)},$$
where $[\boldsymbol{x}]_d\in \re^{C^{d}_{z+d}}$ is the vector of all monomials in $\boldsymbol{x}$ whose degrees are not greater than $d$; ${\rm coef}(f)$ is the vector of coefficients of $f$ corresponding to each monomial in $[\boldsymbol{x}]_d$;
and $\deg(f)$ denotes the degree of polynomial $f$. For example, a polynomial $f_0 = 1 + 2 x_1+3 x_2 + 4x^2_1 + 5x_1 x_2 + 6 x^2_2 $ in $\boldsymbol{x} = (x_1,x_2)$ can be rewritten as $f_0 = (1,2,3,4,5,6)^\top(\boldsymbol{x}^{(0,0)},\boldsymbol{x}^{(1,0)},\boldsymbol{x}^{(0,1)},\boldsymbol{x}^{(2,0)},\boldsymbol{x}^{(1,1)},\boldsymbol{x}^{(0,2)})$. 

A polynomial in $x$ is said to be an Sum-of-Squares (SOS) if it can be written as the sum of squares of other polynomials.
Denote the set of all SOS polynomials as $\Sigma[\boldsymbol{x}]$, then every SOS polynomial $\sigma \in \Sigma[\boldsymbol{x}]\cap \re[\boldsymbol{x}]_{2d}$ can be written in the quadratic form \cite{nie2023moment}:
\begin{equation}\label{eq:quadratic_form}
    \sigma = [\boldsymbol{x}]_d^{\top} X [\boldsymbol{x}]_d,
\end{equation}
where $X\in \mathbf{S}^{\ell_{d}}_+$ is a symmetric positive semidefinite matrix of dimension $\ell_{d} = C^{d}_{z+d} = \left(\begin{array}{c}z+d \\ d\end{array}\right)$. 

We will use $\boldsymbol{x}$ as the 3D coordinates in Euclidean space $\boldsymbol{x}=(x,y,z)$ in the remaining of this paper to describe the geometrical sets in $\re^3$ unless otherwise specified. 
The space occupied by robot $\mathcal{B}$ in its body frame $\{b\}$ is defined with $m$ polynomial inequalities:
\begin{equation*}
   {}^{b}\mathcal{W}_\mathcal{B}:= \{ \boldsymbol{x}\in\re^3: f_1(\boldsymbol{x})\ge0,f_2(\boldsymbol{x})\ge0,\ldots, f_m(\boldsymbol{x})\ge0 \}.
\end{equation*}
Note that the semialgebraic sets expression can represent robots with the majority of convex shapes and part of nonconvex shapes in $\re^3$. 

For a polytopic free region $\mathcal{Q}_{\tau}$ consisting of $r$ facets, we define it by stacking the linear constraints for each hyperplane into $F\in\re^{r\times3}$ and $g\in\re^r$ in its own frame $q$. In this sense, the uniformly scaled free region from $\{q\}$ can be represented as: 
\begin{equation*}
       {}^q\mathcal{Q}_{\tau}(\alpha):= \{ \boldsymbol{x}\in\re^3: (\alpha g^{\top},-F^{\top})^{\top}[\boldsymbol{x}]_1\ge\boldsymbol{0}\},
\end{equation*}
where $\alpha$ simply functions as a ratio on the constant term $g$.
\subsubsection{Certified Containment Relationship with Minimum Scaling}
Since the containment relationship in (\ref{eq:safety_constraints}) can be stated in any reference frame, we formulate the optimization problem in robot body frame $\{b\}$, in which the set ${}^{b}\mathcal{W}_\mathcal{B}$ will not change with the robot state $\boldsymbol{q}$. Denote $R(\boldsymbol{q}) \in \re^{3\times3}$ and $p(\boldsymbol{q})\in\re^3$ the rotation matrix and translation vector representing the configuration of robot body frame $\{b\}$ in $\{q\}$, the scaled free region can be represented in $\{b\}$ as:
\begin{equation*}
       {}^b\mathcal{Q}_{\tau}(\alpha,\boldsymbol{q}):= \{ \boldsymbol{x}\in\re^3: \big((\alpha g-Fp)^{\top},-(FR)^{\top}\big)^{\top}[\boldsymbol{x}]_1\ge\boldsymbol{0}\},
\end{equation*}
where $R$ and $p$ depend on the robot state $\boldsymbol{q}$.
In this subsection, we construct the optimization problem solving for the minimal scaling factor $\alpha$ that certifies the following containment relationship in robot body frame:
\begin{equation}\label{eq:safety_constraints_body_frame}
{}^b\mathcal{W}_\mathcal{B}\subseteq {}^b\mathcal{Q}_{\tau}\left(\alpha,\boldsymbol{q}\right).
\end{equation}
 Let $\mathcal{P}({}^b\mathcal{W}_\mathcal{B})$ be the set of all the \textbf{nonnegative polynomials} on ${}^b\mathcal{W}_\mathcal{B}$, which is defined as:
\[ \mathcal{P}({}^b\mathcal{W}_\mathcal{B}) = \{ f\in \re[\boldsymbol{x}]: f(\boldsymbol{x})\ge0 \ \forall \,\boldsymbol{x}\in {}^b\mathcal{W}_\mathcal{B} \}. \] Then, the minimal scaling problem certifying the set containment relationship (\ref{eq:safety_constraints_body_frame}) at given $\boldsymbol{q}$ is then re-stated in terms of the nonnegative polynomials cone on ${}^b\mathcal{W}_\mathcal{B}$:
\begin{align}\label{eq:safety_constraint_cone_programming}
\displaystyle   \min _{\alpha}  \quad &\alpha \notag
\\ \text { s.t.}  \quad&  f_i^{\mathcal{Q}_{\tau}}(\alpha) \in \mathcal{P}({}^b\mathcal{W}_\mathcal{B})\\
&\quad\quad\forall\,\,i = 1,2,\ldots,r\notag\\
& \alpha\geq0\notag.
\end{align}
Here, $f^{\mathcal{Q}_{\tau}}_i(\alpha)$ represents the $i$th defining linear constraints in ${}^b\mathcal{Q}_{\tau}(\alpha,\boldsymbol{q})$.
Certifying the nonnegativity of a polynomial is generally intractable; therefore, we further introduce the quadratic module (Qmod) given by $f$:
\begin{equation*}
             \qm[f] := \{ \sigma_0 + f_1 \sigma_1 + \ldots + f_{m}\sigma_m: \sigma_0,\sigma_1,\ldots,\sigma_{m} \in \Sigma[\boldsymbol{x}]\}.
\end{equation*}

\begin{theorem}\textup{\cite{putinar1993positive}}
    Suppose $^b\mathcal{W}_B$ is \textbf{Archimedean}\footnote{The set $^b\mathcal{W}_B$ is said to be Archimedean if there exists $\rho\in \qm[f]$ such that the set given by $\rho(x)\ge 0$ is compact. Note that if $^b\mathcal{W}_B$ is compact, then one can always choose a sufficiently large number $N$ and add a redundant ball constraint $N-\sum_{i=1}^n x_i^2\geq0$ to $^b\mathcal{W}_B$ such that the archimedean condition holds.}. If a polynomial $p(\boldsymbol{x})>0$ for all $\boldsymbol{x}\in {^b\mathcal{W}_B}$, then $p\in \textup{Qmod}[f]$. 
   \label{theorem:putinar}
\end{theorem}
The above theorem is known as \textbf{\textit{Putinar's Positivstellensatz}}, which guarantees that all positive polynomials on $P(^b\mathcal{W}_B)$ belong to $\qm[f]$.
Moreover, let the $2k$ truncation of $\qm$, denoted as $\qm[f]_{2k}$, be the subset of $\qm$ where the degree of each element is no greater than $2k$, i.e., $\sigma_0 \in\Sigma[\boldsymbol{x}]\cap\re[\boldsymbol{x}]_{2k}$ and $\sigma_i \in\Sigma[\boldsymbol{x}]\cap\re[\boldsymbol{x}]_{2\lfloor k-\deg(f_i)/2\rfloor}$ for all $i=1,2,\dots,m$.
Then, for all $p\in P(^b\mathcal{W}_B)$ and $\varepsilon>0$ ($\varepsilon$ can be $0$ if $p>0$ on $P(^b\mathcal{W}_B)$), there exists a degree $k$ such that $p+\varepsilon\in \qm[f]_{2k}$. 
We refer to \cite{nie2007complexity,baldi2023effective} for more details on the convergence rate of Putinar's Positivstellensatz.

Consider a nonnegative relaxation order $k$ with $k\geq\lfloor\max\{\deg(f^{\mathcal{Q}_{\tau}}),\deg(f_1)/2,\deg(f_2)/2,\ldots,\deg(f_m)/2\}\rfloor$, the $k$th SOS relaxation of (\ref{eq:safety_constraint_cone_programming}) is formulated as:
\begin{align}\label{eq:minimal_scaling}
\displaystyle   \min _{\alpha, \sigma_{i,j}}  \quad &\alpha 
\\ \text { s.t.}  \quad&  f^{\mathcal{Q}_{\tau}}_i(\alpha) = \sigma_{i,0}+\sum_{j=1}^m f_j \cdot\sigma_{i,j},\tag{\ref{eq:minimal_scaling}{a}}\\
&\quad\quad\forall\,\,i = 1,2,\ldots,r\notag\\
&\sigma_{i,0}\in\Sigma[\boldsymbol{x}]\cap \re[\boldsymbol{x}]_{2k},\tag{\ref{eq:minimal_scaling}{b}}\\
&\sigma_{i,j}\in\Sigma[\boldsymbol{x}]\cap \re[\boldsymbol{x}]_{2\lfloor k-\deg(f_j)/2\rfloor},\tag{\ref{eq:minimal_scaling}{c}}\\
&\quad\quad\forall\,\, i = 1,2,\ldots,r,\,\forall\,\,j=1,2,\dots,m\notag\\
& \alpha\geq0\tag{\ref{eq:minimal_scaling}{d}}.
\end{align}
Given a robot state $\boldsymbol{q}$, specifically the fixed $R(\boldsymbol{q})$ and $p(\boldsymbol{q})$, the formulated problem (\ref{eq:minimal_scaling}) is a standard SOS program since the decision variable $\alpha$ is linear in the coefficients of certain monomials. To facilitate the employment of conic programming solvers and the further derivation of gradient, we transform (\ref{eq:minimal_scaling}) into an SDP by expressing the SOS polynomials into the quadratic form as in (\ref{eq:quadratic_form}): 
\begin{align}\label{eq:minimal_scaling_sdp}
\displaystyle   \min _{\alpha, X_{i,j}}  \quad &\alpha 
\\ \text { s.t.}  \quad&  {\rm coef}(f^{\mathcal{Q}_{\tau}}_i(\alpha))_l - \sum_{j=0}^m\langle A_{(i,l),j},X_{i,j}\rangle\tag{\ref{eq:minimal_scaling_sdp}{a}}=0,\\
&\quad\quad\forall\,\,i = 1,2,\ldots,r,\,\forall\,\,l = 1,2,\ldots,\ell\notag\\
&X_{i,0}\in \mathbf{S}^{\ell_0}_+,\,\ell_0 = \left(\begin{array}{c}z+k \\ k\end{array}\right),\tag{\ref{eq:minimal_scaling_sdp}{b}}\\
&X_{i,j}\in \mathbf{S}^{\ell_j}_+,\,\ell_j = \left(\begin{array}{c}z+\lfloor k-\deg(f_j)/2\rfloor \\ \lfloor k-\deg(f_j)/2\rfloor\end{array}\right),\tag{\ref{eq:minimal_scaling_sdp}{c}}\\
&\quad\quad\forall\,\, i = 1,2,\ldots,r,\,\forall\,\,j=1,2,\dots,m\notag\\
& \alpha\geq0\tag{\ref{eq:minimal_scaling_sdp}{d}}.
\end{align}
In (\ref{eq:minimal_scaling_sdp}), $\langle\cdot,\cdot\rangle$ represents the inner product between two matrices, and $z$ is the dimension of the intermediates $\boldsymbol{x}$. Note that (\ref{eq:minimal_scaling_sdp}a) is obtained by enforcing equality constraints on the coefficients corresponding to all $\ell$ monomials involved in (\ref{eq:minimal_scaling}a).

\Remark{Under mild conditions, optimal solutions of (\ref{eq:safety_constraint_cone_programming}) and (\ref{eq:minimal_scaling_sdp}) are identical for all $k$ that is sufficiently large \cite{nie2014optimality}. Indeed, they are equal for all $k$ if we further assume that all the polynomial functions involved are SOS-convex \cite{nie2023moment}. Therefore, we can determine the minimum scaling factor $\alpha$ in (\ref{eq:safety_constraint_cone_programming}) by solving the SDP (\ref{eq:minimal_scaling_sdp}).}
\subsection{Collision-Free Trajectory Optimization with Implicit Safety Constraints
}
With the geometrical interpretation of the containment relationship in (\ref{eq:safety_constraints}), we define a uniformly scaling factor $\alpha_{\tau}$ at each waypoint $\boldsymbol{q}_{\tau}$ in $\mathcal{X}$, and reformulate the collision-free trajectory optimization problem (\ref{eqn:to}) as:
\begin{align}\label{eqn:to_reformulate}
\displaystyle  \operatorname*{minimize}_{(\boldsymbol{q}_{\tau},\boldsymbol{u}_{\tau})\in\re^{n}\times\re^{m}}\quad & \phi_T(\boldsymbol{q}_{T})+\sum_{\tau=0}^{T-1}J_\tau\big(\boldsymbol{q}_{\tau},\boldsymbol{u}_{\tau}\big)\notag\\ 
    \operatorname*{subject\ to}\quad\,\,  & \boldsymbol{q}_{\tau+1}=f\big(\boldsymbol{q}_{\tau},\boldsymbol{u}_{\tau}\big),\notag\\    & h_{\tau}(\boldsymbol{q}_{\tau},\boldsymbol{u}_{\tau}) = 0,\notag\\ &g_{\tau}(\boldsymbol{q}_{\tau},\boldsymbol{u}_{\tau}) \leq 0,\yesnumber\\
    &\quad\quad \tau = 0,1,\ldots,T-1\notag\\
    & \alpha_{\tau}(\boldsymbol{q_{\tau}})\leq1,\notag\\
    &\quad\quad \tau = 0,1,\ldots,T\notag\\
    &\boldsymbol{q}_0 = \boldsymbol{q}_{s}.\notag
\end{align}
In (\ref{eqn:to_reformulate}), motion safety considering geometry of the robot is ensured by enforcing $\alpha_{\tau}$ to be no greater than one over the entire optimization horizon, where $\alpha_{\tau}(\boldsymbol{q}_{\tau})$ is the optimum of the proposed minimum scaling problem (\ref{eq:safety_constraint_cone_programming}), which is implicitly dependent on the robot configuration $\boldsymbol{q}$ at time stamp $\tau$. 

In this section, to handle the implicit safety constraints, 
we calculate a direction for updating the robot configuration $q$ to decrease the value function $\nu(\boldsymbol{q})$ of (\ref{eq:minimal_scaling_sdp}), i.e., the minimum scaling factor $\alpha$ in (\ref{eq:minimal_scaling_sdp}) with the given robot configuration $q$. 
In practical implementation, we use $-\partial {\mathcal{L}}/\partial {\boldsymbol{q}}$ at the primal-dual solution of (\ref{eq:minimal_scaling_sdp}) to update $\boldsymbol{q}$, which performs greatly in computation and has also been widely exploited for trajectory optimization in robotics \cite{10160716,10184036}.
Indeed, one may show that under mild conditions, the gradient $\partial {\mathcal{L}}/\partial {\boldsymbol{q}}$ belongs to an upper approximation for the Clarke subgradient of $\nu(\boldsymbol{q})$; see \cite[Corollary~1, page~242]{clarke1990optimization}.
Furthermore, if we additionally suppose $\nu(\boldsymbol{q})$ is differentiable (this is the case when (\ref{eq:minimal_scaling_sdp}) has a unique primal-dual solution and Slater's condition holds), then $\nabla \nu = \partial {\mathcal{L}}/\partial {\boldsymbol{q}}$, by \cite[Corollary~2, page~242]{clarke1990optimization}.


In (\ref{eq:minimal_scaling_sdp}), since the rotation matrix $R(\boldsymbol{q})$ and translation vector $p(\boldsymbol{q})$ appear in the coefficients ${\rm coef}(f^{\mathcal{Q}_{\tau}}_i(\alpha))_l$, it is only necessary to consider the terms in (\ref{eq:minimal_scaling_sdp}a) when computing the gradient $\partial\mathcal{L}/\partial\boldsymbol{q}$. Therefore, the gradient vector can be represented as:
\begin{align}\label{eq:gradient}
     \frac{\partial\mathcal{L}}{\partial\boldsymbol{q}}
     = \sum_{i,l}\lambda_{i,l}\frac{\partial{\big({\rm coef}(f^{\mathcal{Q}_{\tau}}_i)_l}\big)}{\partial\boldsymbol{q}},
\end{align}
where $\lambda_{i,l}$ is the dual variable for each constraint in (\ref{eq:minimal_scaling_sdp}a). It is worthwhile to highlight that, with the analytical solution in (\ref{eq:gradient}), we can efficiently determine the gradient value after solving (\ref{eq:minimal_scaling_sdp}) to its optimum.


Subsequently, we employ the gradient-based algorithm Augmented Lagrangian Iterative Linear Quadratic Regulator (AL-iLQR) proposed in \cite{8967788}. In this approach, the constraints can be augmented into the objective function and solved iteratively using the iLQR algorithm. Specifically, we calculate the value of $\alpha$ as well as the gradient information based on the trajectory $\boldsymbol{q}$ output from each round of forward pass of AL-iLQR. We then integrate this information in the execution of the backward pass of AL-iLQR to calculate the perturbed action, forcing the robot to stay inside the free regions $\mathcal{Q}$. 
To this end, the proposed collision-free trajectory optimization method considering specific robot geometries is summarized in Algorithm \ref{alg:cfto}. 

\Remark{In the proposed bi-level optimization process for solving (\ref{eqn:to_reformulate}), safety is ensured by enforcing the minimal scaling factor $\alpha$ to be less than 1 along the trajectory and using this as a termination criterion for the AL-iLQR to converge.}

\normalem
\begin{algorithm}[t]\small
\caption{Collision-Free Trajectory Optimization with SOS Programming}
\label{alg:cfto}
\SetKwFunction{MyFunction}{WaypointsAllocation}
\SetKwProg{Fn}{Function}{:}{}
\SetKwFor{For}{parfor}{:}{end}
\KwIn{$\mathcal{Q}$, $\boldsymbol{q}_s$, $\boldsymbol{q}_g$, $T$, $ctol$, and ${}^b\mathcal{W}_\mathcal{B}$}
\KwOut{$\mathcal{X}^*$ and $U^*$}
\DontPrintSemicolon
\Fn{\MyFunction{$\mathcal{Q}, \boldsymbol{q}_s, \boldsymbol{q}_g, T$}}{
    $\mathcal{Q}\gets${Approximate $\mathcal{F}$ with sequence of free regions};\;
    $\mathcal{V} \gets$ {Assign each region $\mathcal{Q}_i$ in $\mathcal{Q}$ to a vertex $v_i$}; \;
    $\mathcal{E} \gets$ Connect ($v_i$,$v_j$) in $\mathcal{V}$ if they are overlapped; \;
    $\mathbb{P}^*_v \gets$ Search for free regions sequence connecting $\boldsymbol{q}_s$ and $\boldsymbol{q}_g$ with shortest path length; \;
    $\mathbb{P}^*_e \gets$ Extract the edge sequence concatenating vertexes in $\mathbb{P}^*_v$; \;
    $\texttt{Index} \gets$ Assign each waypoint to a free region with $\mathbb{P}^*_e$ and the optimization horizon $T$; \;
    $\texttt{Index} \gets$ Add additional regions and reassign waypoints; \;
    \KwRet{$\texttt{Index}$}
}
\textbf{Initialization:}\; $\mathcal{X}^* \gets \mathcal{X}_0$ and $U^*\gets U_0$;\;$ctol$ $\gets$ Tolerance of constraints violation in AL-iLQR;\;\texttt{c} $\gets$ Constraints violation of current trajectory;\;
\texttt{Index} $\gets$ \MyFunction{$\mathcal{Q}, \boldsymbol{q}_s, \boldsymbol{q}_g, T$};\; 
\While {$\texttt{c} \ge ctol$}{
\For{$\boldsymbol{q}_{\tau}\in \mathcal{X}^*$}{
$\alpha_{\tau}, {\partial\alpha_{\tau}}/{\partial\boldsymbol{q}} \gets $ Solve (\ref{eq:minimal_scaling_sdp}) and calculate gradients from (\ref{eq:gradient}) with \texttt{Index}; }$\delta u \gets$ \texttt{ALiLQR-BACKWARDPASS}($\mathcal{X}^*$,\,$U^*$, $\alpha$,\,$\frac{\partial\alpha}{\partial\boldsymbol{q}}$);\;
$\mathcal{X}^*,U^*,\texttt{c}\gets$\texttt{ALiLQR-FORWARDPASS}($\mathcal{X}^*,\,U^*,\,\delta u$)

}

\end{algorithm}
\ULforem

\section{Results}
In this section, we thoroughly evaluate our proposed trajectory optimization approach through simulations and real-world experiments. For simulations, we run the algorithm on an Intel i5-13400F processor, and the proposed SDP in (\ref{eq:minimal_scaling_sdp}) is implemented and solved parallelly using MOSEK \cite{mosek} with precomputation of the pertinent parameters. Initially, we validate our algorithm on robots modeled with various geometries represented as semialgebraic sets. We then demonstrate the efficacy and robustness of our algorithm for collision-free trajectory optimization in three challenging maze scenarios with varying levels of complexity. Also, we conduct comparative experiments to further showcase the safety and computational advantages of our method in complex and obstacle-dense environments. Finally, we apply the trajectory optimization pipeline on the real robot platform Unitree Go1, with the entire process implemented on an intel NUC11 with i5-7260U processor, highlighting the adaptability and practicality of our method for safely navigating through a cluttered indoor environment.

\subsection{Simulations}
\subsubsection{Algorithm Validation for Various Robot Geometries}
\begin{figure}[t]	
	\centering
 	\includegraphics[width=0.9\linewidth]{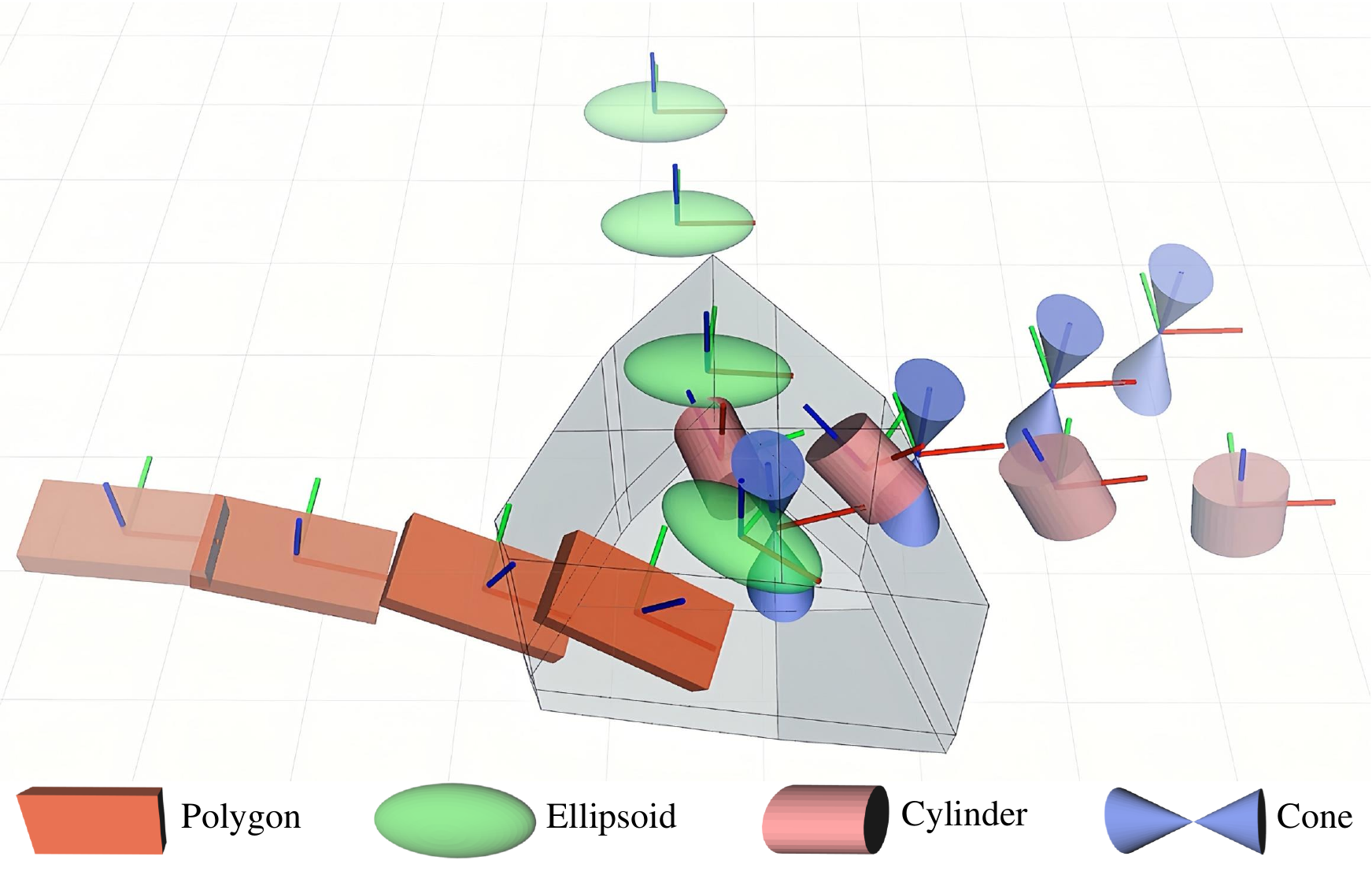}

	\caption
	{Visualization of the optimized trajectories for robots with different geometries. The robots are commanded to move into the free region (shown in grey) from random start configurations by enforcing our derived safety constraints at the terminal time stamp.}
	\label{fig:converge_ill}\vspace{-1em}
\end{figure}

In this scenario, we validate the effectiveness of our proposed safety constraints on robots with different geometries represented as semi-algebraic sets, including a polytope, a cylinder, an ellipsoid, and an elliptical cone. We initialize the trajectories at random configurations and command the robots to stop within the safe region by enforcing the terminal safety constraint, i.e., $\alpha_{T}(\boldsymbol{q}_{T}) \leq 1$. As shown in Fig. \ref{fig:converge_ill}, trajectories for robots with all geometries have successfully been ``attracted" into the designated region utilizing the proposed bi-level optimization framework with the scaling-based safety constraints. We achieve 10\,$\pm$\,5\,$\textup{ms}$ in average for solving the proposed SDP at the minimum relaxation order since the polynomial functions defining the geometries are all SOS-convex, and the time spent on formulating the SDP and calculating the gradient can be neglected as the explicit expressions have been pre-computed.

\begin{figure}[t]
    \centering
	\includegraphics[width =\linewidth]{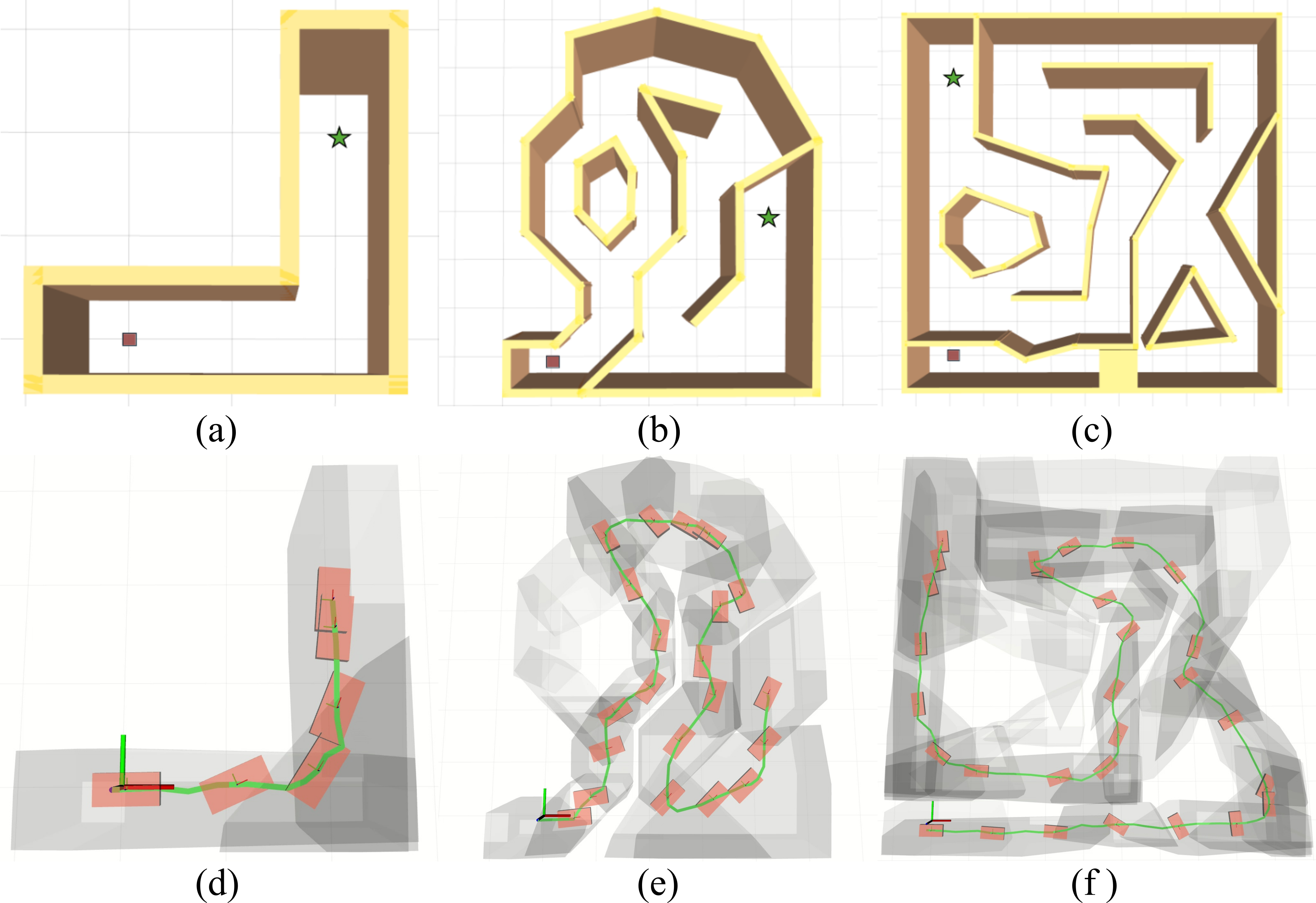}

	\caption
	{Performance of our proposed trajectory optimization framework in maze traversing tasks. (a-c) Three maze scenarios with different degrees of complexity. The start and goal positions are denoted by red and green nodes, respectively. (d-f) Visualization of the collision-free trajectories generated in the graph of free regions $\mathcal{G}$ with keyframes highlighted.}
     \label{fig:maze}\vspace{-1em}
\end{figure}

\begin{figure*}[t]%
    \centering
    \subfloat[Kinodynamic RRT*\cite{6631299}]{
        \includegraphics[width=0.16\linewidth]{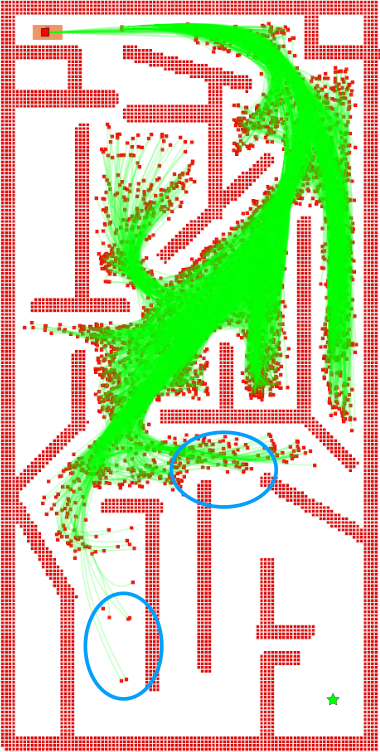}
        }\hfill
    \subfloat[TrajOpt\cite{doi:10.1177/0278364914528132}]{
        \includegraphics[width=0.16\linewidth]{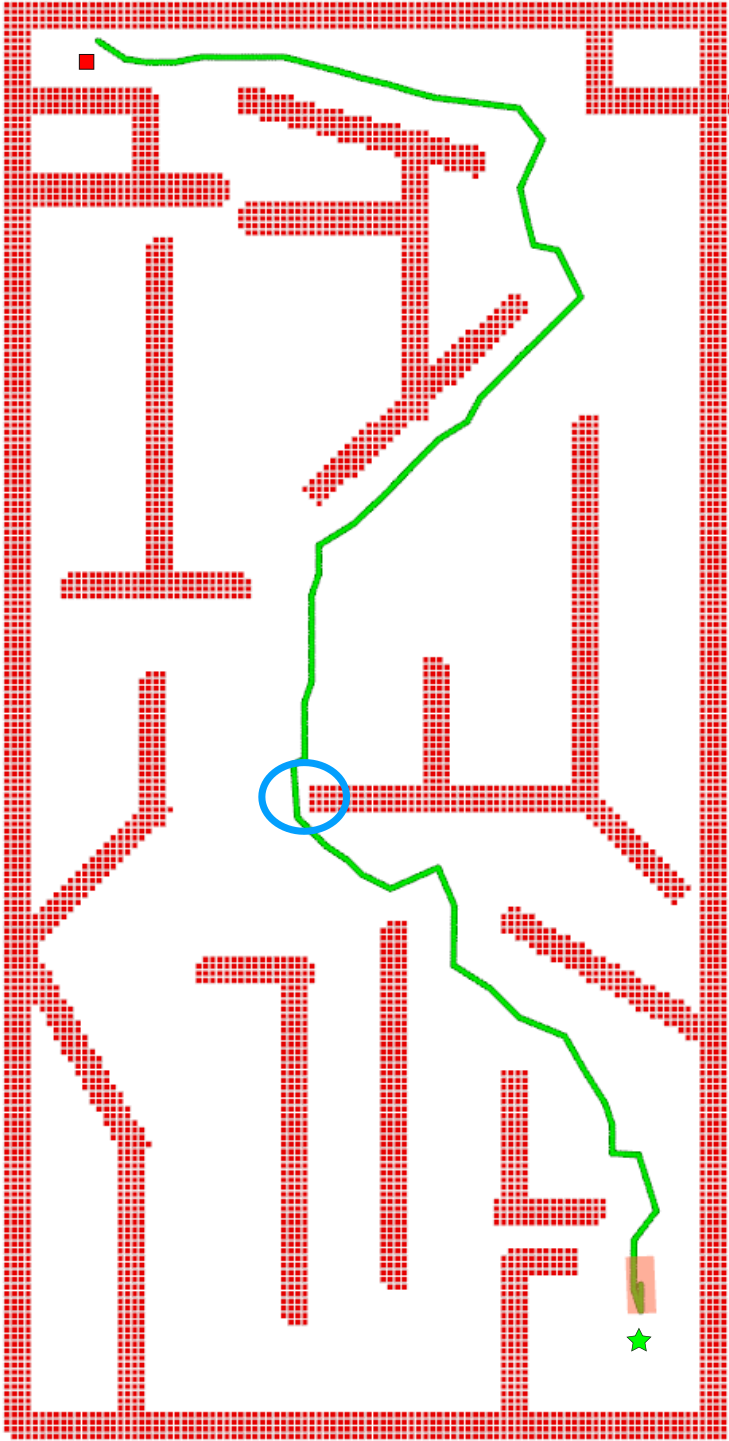}
        }\hfill
    \subfloat[CBF-Polytopes\cite{9812334}]{
        \includegraphics[width=0.16\linewidth]{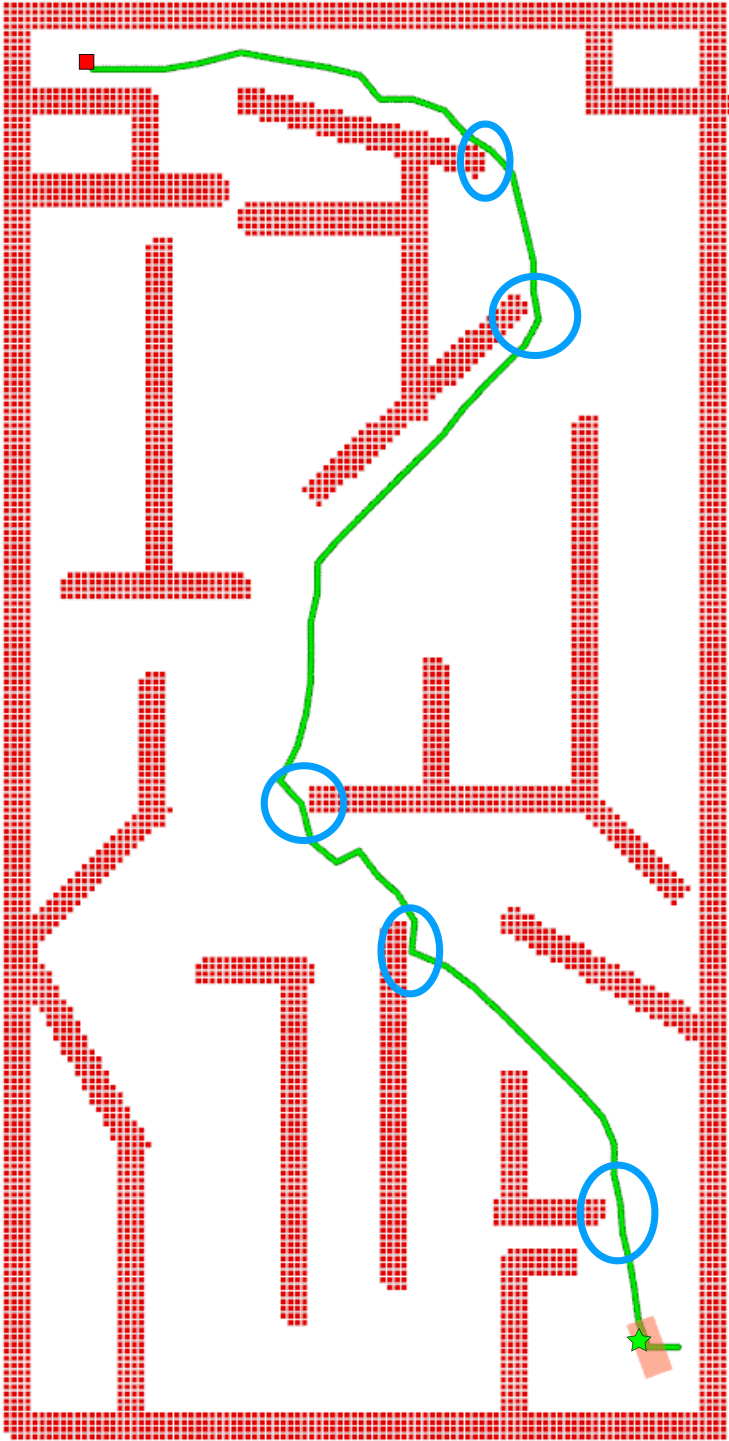}
        }\hfill
    \subfloat[Safe-Corridor\cite{7839930}]{
        \includegraphics[width=0.16\linewidth]{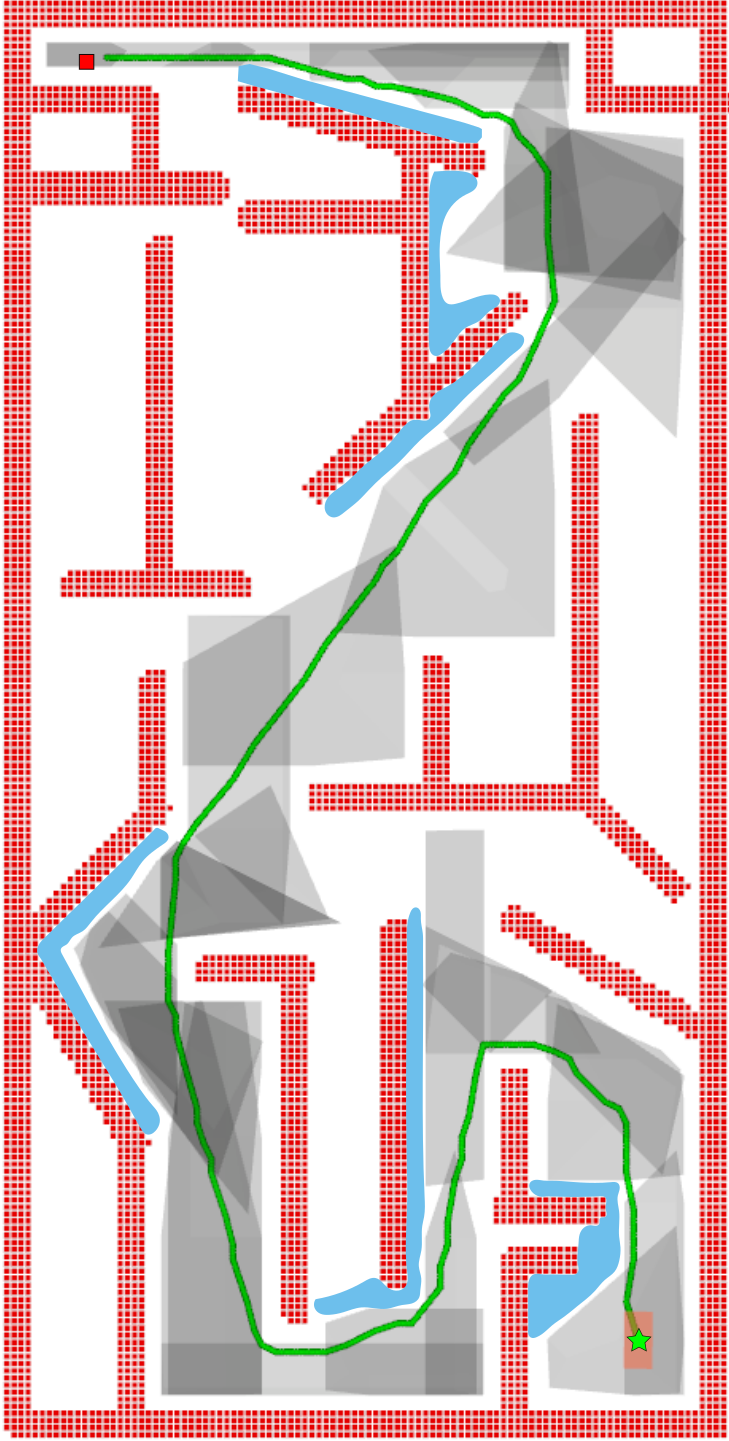}
        }\hfill
    \subfloat[Ours]{
        \includegraphics[width=0.16\linewidth]{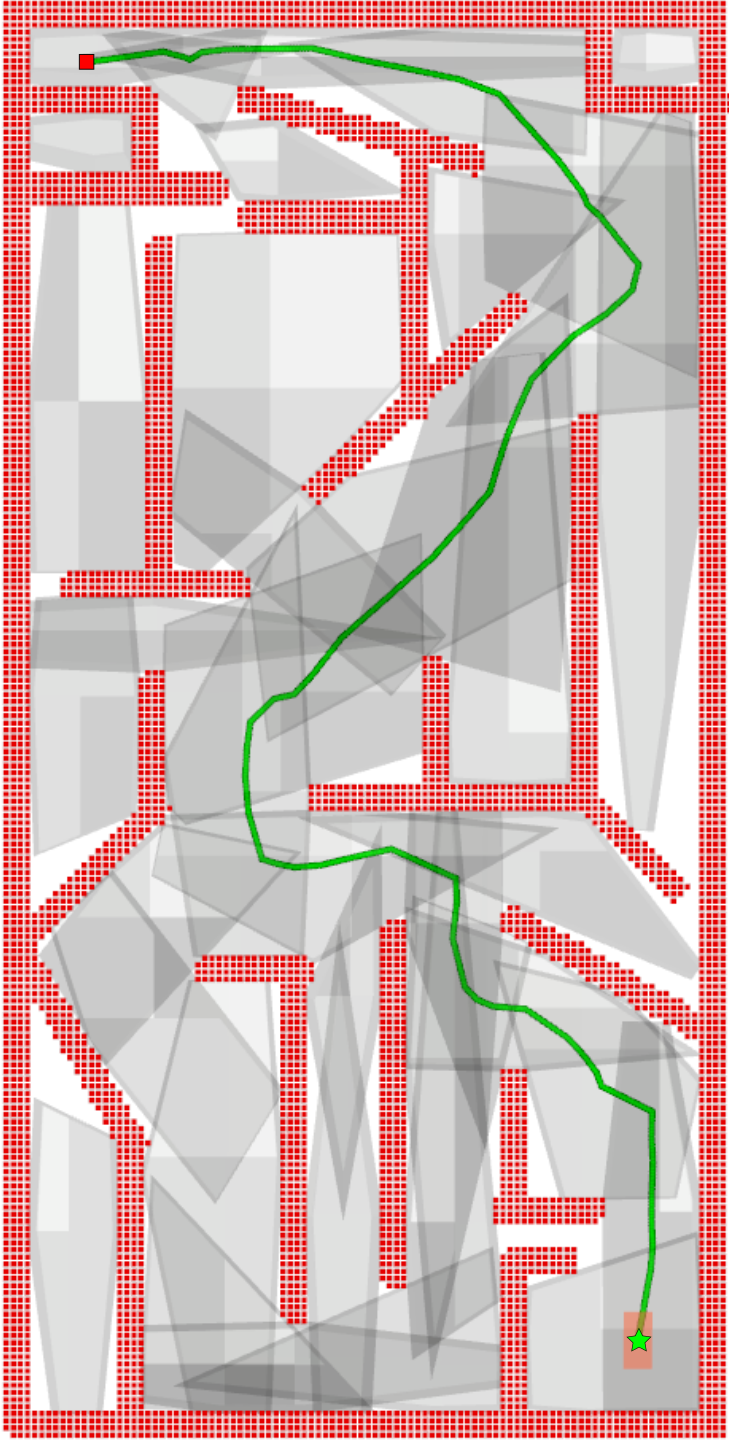}
        }
    \caption{Comparsion results in a 5\,\textup{m}$\times$10\,\textup{m} cluttered environment. A 0.2\,\textup{m} $\times$ 0.4\,\textup{m} polytopic robot is commanded to move from initial (red node) to goal configuration (green node). (a) Visualization of the search process of Kinodynamic RRT*. It fails to find a feasible trajectory through the narrow passages (highlighted) within tractable time (100 \textup{s}). (b-c) Optimized trajectories of TrajOpt and CBF-Polytopes. They fail to push the robot away from all the obstacles simultaneously within the optimization horizon, and the collision areas are highlighted. (d-e) Visualization of the generated free regions and the planned trajectories of Safe-Corridor and our proposed method. Both approaches generate safe trajectories by decomposing free space. However, Safe-Corridor searches for paths and extracts free regions based on obstacle information that is dilated by the robot's maximum side size. This method fails to find the nearest traversable path. Additionally, as highlighted by the gaps between the generated free regions and the obstacles in (d), the free regions do not cover enough of the available free space, resulting in a conservative trajectory.}
    \label{fig:exp_comparison}
    \vspace{-1em}
\end{figure*}

\subsubsection{Traversing in Mazes}
To evaluate the performance of our proposed trajectory optimization framework in challenging scenarios, we have crafted three distinct mazes of varying dimensions and degrees of complexity, namely the small, medium, and large mazes, respectively, as shown in Fig. \ref{fig:maze}(a)-\ref{fig:maze}(c). Remarkably, our method successfully optimizes collision-free trajectories in all three scenarios with fast convergence, taking 0.48\,\textup{s}, 2.5\,\textup{s}, and 4.2\,\textup{s}, respectively. As depicted in Fig. \ref{fig:maze}(d)-\ref{fig:maze}(f), our optimized trajectories enable the robot to navigate through diverse complex maze environments by adopting different degrees of posture adjustments based on its shape, thereby demonstrating the effectiveness of our collision-free trajectory optimization method.  
\label{sec:forest}

\begin{table}[t]
 \centering
  \caption{Comparisons between our method and benchmark methods in a cluttered maze environment}
\resizebox{\columnwidth}{!}{%
\begin{tabular}{cccccc}
\toprule[1pt]
\multirow{2}{*}{\large\textbf{Methods}} &
  \multirow{2}{*}{\large\textbf{\begin{tabular}[c]{@{}c@{}}Trajectory\\  Length (m)\end{tabular}}} &
  \multicolumn{2}{c}{\large\textbf{Computation Time (s)}}  &  \multirow{2}{*}{\large\textbf{Safety}} &
  \multirow{2}{*}{\large\textbf{Non-conservative}} \\
   \cmidrule(lr){3-4}
                 &       &\ \large\textbf{Map Processing} & \large\textbf{Optimization} &     \\ \specialrule{0em}{2pt}{2pt}\hline \specialrule{0em}{2pt}{2pt}
\large Kinodynamic RRT*\cite{6631299} & -     & -                       & \large 100+               & \large\textcolor{red}{$\times$}     & \large\textcolor{red}{$\times$}  \\ \specialrule{0em}{1pt}{1pt}
\large TrajOpt\cite{doi:10.1177/0278364914528132}          & \large 13.95 & \large 0.013                   & \large 6.925              & \large\textcolor{red}{$\times$}     & \large\textcolor{green}{\checkmark}  \\\specialrule{0em}{1pt}{1pt}
\large CBF-Polytopes\cite{9812334}     & \large 13.45 & \large 0.013                   & \large 33.403             & \large\textcolor{red}{$\times$}     & \large\textcolor{green}{\checkmark}  \\\specialrule{0em}{1pt}{1pt}
\large Safe-Corridor\cite{7839930}    & \large 17.97 & \large 0.018                   & \large 0.256            & \large\textcolor{green}{\checkmark}       & \large\textcolor{red}{$\times$} \\\specialrule{0em}{1pt}{1pt}
\large\textbf{Ours}             & \large\textbf{13.70} & \large0.031                   & \large\textbf{0.561}           & \large\textcolor{green}{\checkmark}      & \large\textcolor{green}{\checkmark} \\ \bottomrule[1pt]
\end{tabular}%
}\label{tab:compare}%
\end{table}
\subsubsection{Comparisons}
We benchmark our proposed framework against Kinodynamic RRT*\cite{6631299}, TrajOpt\cite{doi:10.1177/0278364914528132}, CBF-Polytopes\cite{9812334}, and Safe-Corridor\cite{7839930}. The objective is to plan motions for a robot with polytopic geometry from the initial state to the goal state in a cluttered environment with complex obstacle layouts. Each of these methods differs from our approach in how it handles the environmental (obstacles) information, ensures safety, and solves the problem. For TrajOpt and CBF-Polytopes, we use the jump points search (JPS) algorithm to generate path references with no obstacle dilation, while in Safe-Corridor, obstacles are dilated based on the robot’s maximum side length in searching the reference path. Subsequently, it formulates a minimum snap QP problem and confines waypoints on the trajectory within the free regions extracted along the path. We compare computation time, trajectory length, and safety among the methods, with results recorded in Table \ref{tab:compare}. The trajectories generated by each method are shown in Fig. \ref{fig:exp_comparison}.

As indicated in the Fig. \ref{fig:exp_comparison}(a)-\ref{fig:exp_comparison}(c), Kinodynamic RRT*, TrajOpt, and CBF-Polytopes all failed to optimize safe trajectories. Kinodynamic RRT*, being a sampling-based method, struggles to generate valid paths due to the limited number of samples that fall in constrained spaces, reducing the probability of randomly sampling feasible paths through narrow passages. TrajOpt pushes the robot toward safe areas using locally calculated signed distances from the obstacle information, which can easily lead to getting stuck in local optima when facing complex obstacle layouts\cite{9196996}. In the implementation of CBF-Polytopes, deriving CBF constraints for each obstacle introduces a large number of variables to the optimization problem, reducing computational efficiency, especially in cluttered environments. Additionally, direct methods like SQP and IPOPT used by TrajOpt and CBF-Polytopes are relatively slower compared to the DDP-based method AL-iLQR adopted in our framework, as shown in Table \ref{tab:compare}. On the other hand, both Safe-Corridor and our method generate safe trajectories by exploring the relationship between robots and free regions in cluttered environments. However, Safe-Corridor treats the robot as a point and expands obstacles based on the robot's maximum side length, searching for the reference path using the expanded obstacle information. As demonstrated in Fig. \ref{fig:exp_comparison}, it fails to find the traversable path with the shortest path length, and the trajectory is generated in the shrunk free region (some of the gaps between the free regions and the obstacles are highlighted), leading to a conservative motion.
Additionally, it lacks strict safety guarantees and can only provide a safe trajectory when densely sampled waypoints are included in the safety constraints when solving the backend minimum snap problem.

Our method, however, addresses these limitations by ensuring safety through the bi-level solving process with AL-iLQR. We solve the proposed minimum scaling SDP problem along the trajectory of 64 time stamps in parallel after each forward iteration. Each batch takes approximately 6\,$\pm$\,3\,\textup{ms}, with the entire AL-iLQR converging in 0.561\,\textup{s}. The results show that our method achieves a shorter trajectory length and faster optimization time in cluttered environments compared to the benchmark methods, while also ensuring safety and non-conservativeness.

\begin{figure}[t]
	\centering
        \includegraphics[width=\linewidth]{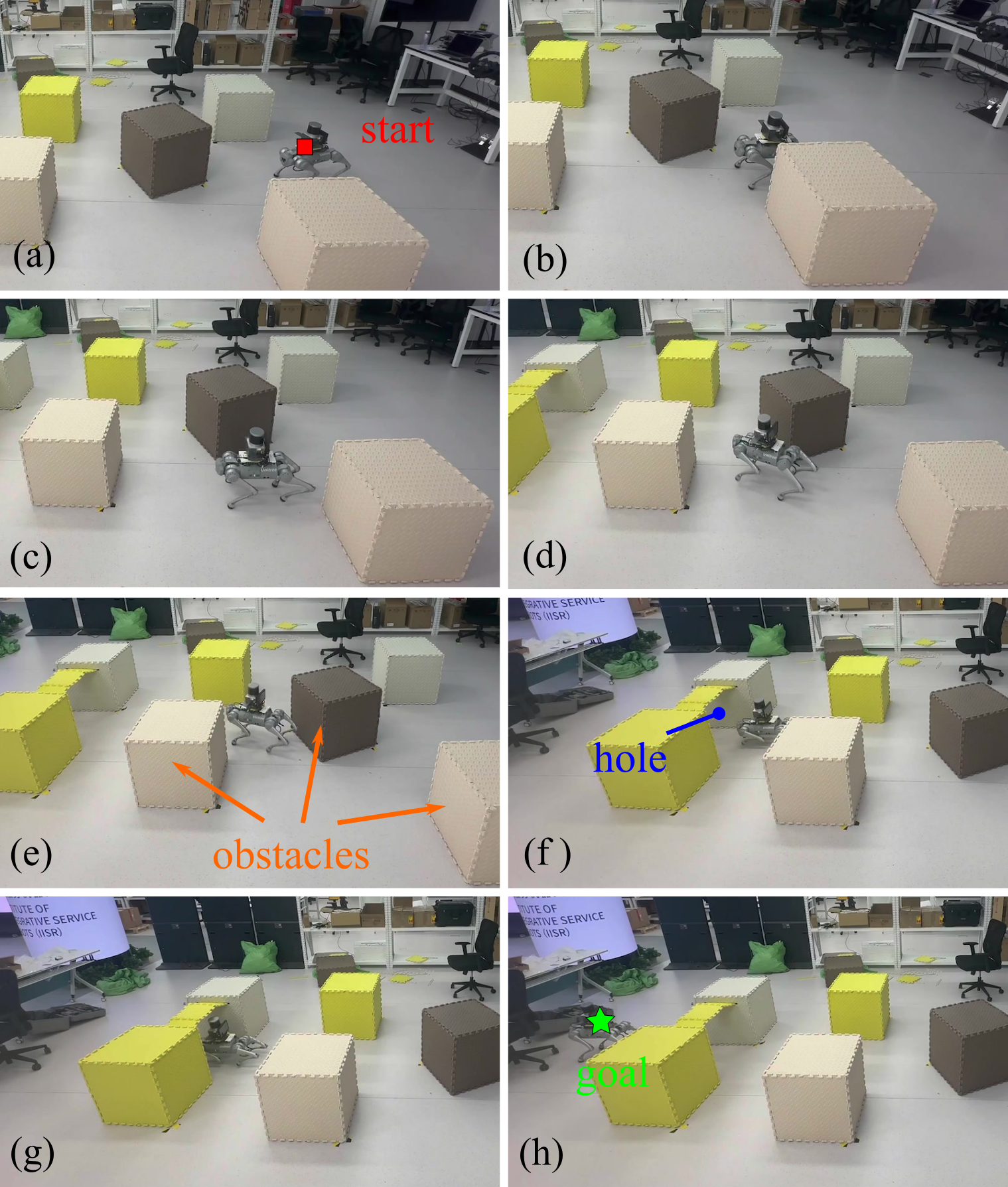}

	\caption
	{Snapshots of real-world experiments with our proposed method. The quadruped is commanded to traverse through the cluttered indoor environment with randomly placed obstacles and a hole.}
	\label{fig:real_exp}\vspace{-1em}
\end{figure}

\subsection{Real-World Experiment}
 In this section, we demonstrate the practicality and robustness of our proposed method by deploying it on a real robot platform Unitree Go1. In the experiment, the quadruped is commanded to traverse through a cluttered indoor area of $2\,\textup{m}\times
6\,\textup{m}$ with several randomly placed obstacles and a hole to simulate the height-constrained situation. 
We use the onboard Robosense 16-line spinning Lidar to perceive point cloud data for off-line space decomposition and real-time align-based localization. 
We incorporate additional physical constraints of the quadruped in the trajectory optimization process and implement the kinematic feedforward and feedback controllers for trajectory tracking. The computed body twist commands are subsequently fed into the embedded speed controller to actuate each joint on the robot. The robot, along with the onboard sensor, is approximated as a polytope. The proposed optimization problem (\ref{eq:minimal_scaling_sdp}) is solved in around 15\,$\pm$\,3$\,\textup{ms}$. As depicted in Fig. \ref{fig:real_exp}, the quadruped adjusts its pose to adapt to the obstacle configurations, and due to the tightly fitted robot geometry, the quadruped lowers its pose to successfully traverse the height-constrained hole in a non-conservative manner (shown in Fig. \ref{fig:real_exp}(g)-\ref{fig:real_exp}(h)). 

\section{CONCLUSION}
This paper introduces a novel framework for optimizing collision-free trajectories to navigate robots safely in cluttered environments. Our approach decomposes the free space into overlapping polytopic regions and formulates an SOS optimization problem to determine the minimum scaling factor for each region, such that the robot with specific geometry is contained within the scaled free region. We enforce the geometry-aware safety constraints by restricting the scaling factor to be less than 1 along the entire trajectory. Gradient information of the SOS optimization problem is then integrated with AL-iLQR to generate collision-free trajectories. We extensively evaluate our proposed algorithm in both simulations and real-world experiments, and the results demonstrate that our method is capable of generating effective and flexible obstacle avoidance maneuvers when navigating the robot through challenging environments.
\bibliographystyle{IEEEtran}
\normalem
\bibliography{ref}

\end{document}